\documentclass[10pt,twocolumn,letterpaper]{article}

\usepackage{cvpr}              

\usepackage[accsupp]{axessibility}

\usepackage[dvipsnames]{xcolor}

\usepackage{multirow}

\definecolor{cvprblue}{rgb}{0.21,0.49,0.74}
\usepackage[pagebackref,breaklinks,colorlinks,citecolor=cvprblue]{hyperref}

\def\paperID{5257} 
\def\confName{CVPR}
\def\confYear{2024}

\title{BIVDiff: A Training-Free Framework for General-Purpose Video Synthesis \\ via Bridging Image and Video Diffusion Models}

\author{Fengyuan Shi$^1$ \quad Jiaxi Gu$^2$ \quad Hang Xu$^2$ \quad Songcen Xu$^2$ \quad Wei Zhang$^2$ \quad Limin Wang$^{1,3}\thanks{}$\\
$^1$State Key Laboratory for Novel Software Technology, Nanjing University \\ $^2$ Huawei Noah's Ark Lab \quad  $^3$ Shanghai AI Laboratory \\
{\tt\small \{fengyuanshi1999, imjiaxi, chromexbjxh\}@gmail.com}
{\tt\small \{xusongcen, wz.zhang\}@huawei.com} \\
{\tt\small lmwang@nju.edu.cn} \\
\textbf{\url{https://bivdiff.github.io}} \\
\vspace{-18mm}
}

\begin{document}

\twocolumn[{
\maketitle
\renewcommand\twocolumn[1][]{#1}
    \begin{center}
        \centering
        \includegraphics[width=1.0\linewidth]{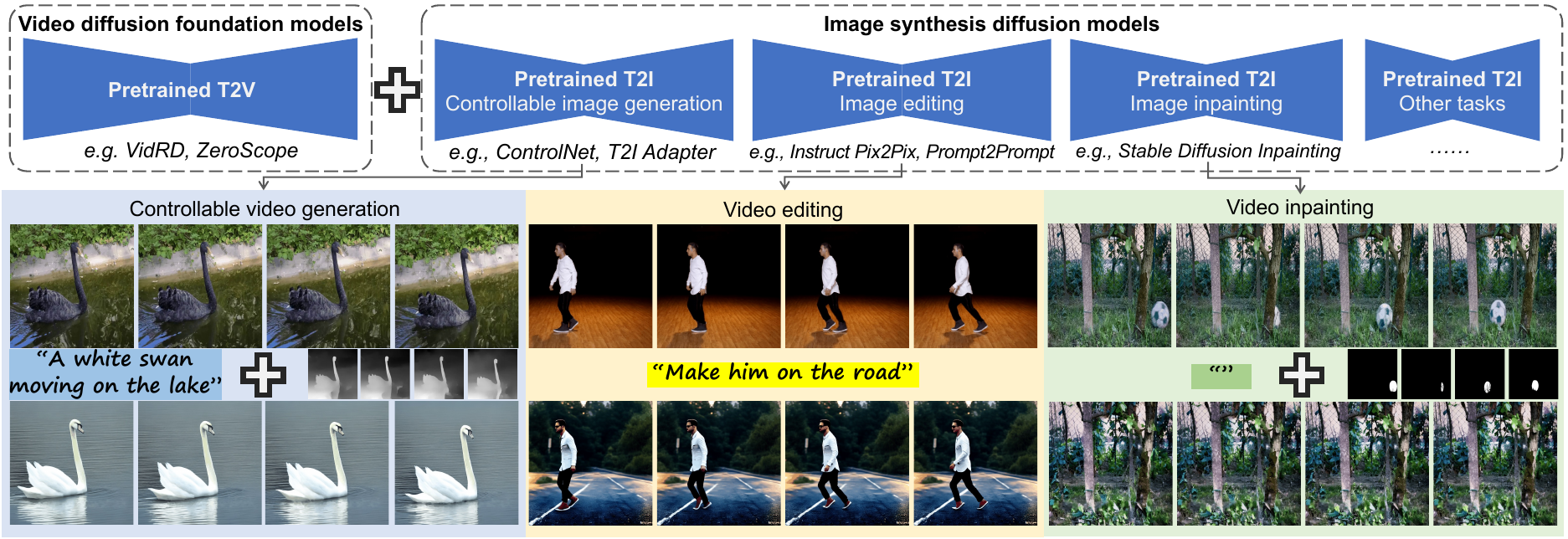}
        \vspace{-6mm}
        \captionof{figure}{Given an image diffusion model (IDM) for a specific image synthesis task, and a text-to-video diffusion foundation model (VDM), our model can perform training-free video synthesis, by bridging IDM and VDM. }
        \label{fig:showcase}
    \end{center}
}]

{
\renewcommand{\thefootnote}
    {\fnsymbol{footnote}}
\footnotetext[1]{Corresponding author.}
}

\begin{abstract}
Diffusion models have made tremendous progress in text-driven image and video generation. Now text-to-image foundation models are widely applied to various downstream image synthesis tasks, such as controllable image generation and image editing, while downstream video synthesis tasks are less explored for several reasons. First, it requires huge memory and computation overhead to train a video generation foundation model. Even with video foundation models, additional costly training is still required for downstream video synthesis tasks. Second, although some works extend image diffusion models into videos in a training-free manner, temporal consistency cannot be well preserved. Finally, these adaption methods are specifically designed for one task and fail to generalize to different tasks. To mitigate these issues, we propose a training-free general-purpose video synthesis framework, coined as {\bf BIVDiff}, via bridging specific image diffusion models and general text-to-video foundation diffusion models. Specifically, we first use a specific image diffusion model (e.g., ControlNet and Instruct Pix2Pix) for frame-wise video generation, then perform Mixed Inversion on the generated video, and finally input the inverted latents into the video diffusion models (e.g., VidRD and ZeroScope) for temporal smoothing. This decoupled framework enables flexible image model selection for different purposes with strong task generalization and high efficiency. To validate the effectiveness and general use of BIVDiff, we perform a wide range of video synthesis tasks, including controllable video generation, video editing, video inpainting, and outpainting. 
\end{abstract}    
\section{Introduction}
\label{sec:introduction}

Diffusion models~\cite{sohl2015deep,ho2020denoising,song2020score} have shown impressive capabilities in generating diverse and photorealistic images. By scaling up dataset and model size, large-scale text-to-image diffusion models~\cite{dhariwal2021diffusion,nichol2021glide,ho2022classifier,ramesh2022hierarchical,saharia2022photorealistic,rombach2022high} gain strong generalization ability and make tremendous breakthroughs in text-to-image generation. By fine-tuning these powerful image generation foundation models on high-quality data in specific areas, various downstream image synthesis tasks also come a long way, such as controllable image generation~\cite{mou2023t2i,zhang2023adding}, image editing~\cite{brooks2023instructpix2pix,DBLP:conf/iclr/HertzMTAPC23,mokady2023null}, personalized image generation~\cite{ruiz2023dreambooth,DBLP:conf/iclr/GalAAPBCC23}, and image inpainting~\cite{rombach2022high}. However, video diffusion models are less explored for different video synthesis tasks due to several critical issues.

First, training video generation foundation models requires substantial training on a massive amount of labeled video data, heavily depending on a large scale of computing resources~\cite{ho2022video,ho2022imagen,DBLP:conf/iclr/SingerPH00ZHYAG23,ge2023preserve,gu2023reuse}. Even with video foundation models available, additional training on high-quality data in specific areas is still required for downstream video synthesis tasks such as controllable video generation~\cite{esser2023structure,wang2023videocomposer} and video editing~\cite{liew2023magicedit,molad2023dreamix}. To improve training efficiency, Tune-A-Video~\cite{wu2023tune} fine-tunes a pre-trained text-to-image model on the input video. Although Tune-A-Video can learn temporal consistency, this kind of per-input fine-tuning is still time-consuming. And it may overfit the small number of input videos and its generalization ability is limited (e.g., poor motion editability). Second, while some works extend image diffusion models into videos in a training-free manner, their temporal consistency cannot be well kept and flickers can still be observed (e.g., \cref{fig:compairison}), due to the weak temporal modeling. Finally, previous works are usually proposed for one specific task and it requires different methods to extend from images to videos for different downstream video synthesis tasks with limited cross-task generality.

Image generation models can exhibit strong generalization and diversity, and yield many powerful downstream image synthesis models through fine-tuning. But frame-wise video generation with image models would lead to temporal inconsistency. Video generation foundation models can generate temporally coherent videos but require additional costly training for downstream video synthesis tasks. A question arises naturally: {\em Is it possible to build a training-free framework for general-purpose video synthesis by jointly leveraging the strengths of both pre-trained image and video diffusion models?} The key challenge is how to design a simple and general interface to bridge these two types of diffusion models to efficiently achieve temporal consistency in video synthesis.

To this end, we propose a general training-free video synthesis framework (BIVDiff), via bridging the {\em specific} image diffusion models and a {\em general} text-to-video diffusion model. Specifically, we first use a task-specific image diffusion model (like ControlNet~\cite{zhang2023adding}, Instruct Pix2Pix~\cite{brooks2023instructpix2pix}) to generate the target video in a frame-by-frame manner, then perform DDIM Inversion~\cite{DBLP:conf/iclr/SongME21} on the generated video, and finally input the inverted latents into the video diffusion model (VDM) for temporal smoothing. Decoupling image and video models enables flexible model selection for different synthesis purposes, which endows the framework with strong task generalization and high efficiency (\cref{fig:showcase}). 

Despite using inverted latents by image DDIM Inversion, VDM tends to generate contents inconsistent with IDM in some cases, due to the distribution shifts. Moreover, for the case with a large gap between the latent distributions of image and video diffusion models, VDMs will fail to generate videos. For example, in the case of inputting source videos, the initial noisy frame latents obtained by frame-wise DDIM Inversion of image diffusion models are highly correlated, making some VDMs (e.g., VidRD~\cite{gu2023reuse}) with i.i.d. random latent requirement collapse to meaningless noises (\cref{fig:ablation_mixing_ratio}). Accordingly, we introduce an improved version called Mixed Inversion. Specifically, we perform DDIM Inversion with both image and video diffusion models. Both latents by Image and Video DDIM Inversion encode the content of videos. The former could be further temporally smoothed by VDM but its distribution may be different from the one required by VDM. The latter cannot be further temporally smoothed by VDM but the distribution is consistent with VDM. We use a weighted sum of these two latents to adjust the distribution of initial latents fed into VDM. With this Mixed Inversion, we can flexibly adjust the latent distribution to make VDMs produce more consistent and better results, and trade off between temporal smoothing and open generation capability of VDMs.  
To validate the effectiveness of BIVDiff, we perform experiments on various representative video synthesis tasks, including 1) Controllable Video Generation; 2) Video Editing; and 3) Video Inpainting and Outpainting. Our contributions are summarized as follows:
\begin{itemize}
    \item We propose a general training-free video synthesis framework, via bridging downstream task-specific image diffusion models and text-to-video diffusion models. Our BIVDiff is simple, efficient, and generalizable for different video synthesis tasks.
    \item We introduce Mixed Inversion, i.e., mixing the DDIM inverted latents of image and video diffusion models, to adjust the latent distribution to make VDMs produce more consistent and better results, and trade off between temporal smoothing and open generation capability of VDMs. 
    \item We perform extensive experiments on various video synthesis tasks, including controllable video generation video editing, video inpainting, and outpainting, demonstrating the effectiveness and general use of BIVDiff.
\end{itemize}

\section{Related Work}
\label{sec:reletaed_work}

\begin{figure*}[t]
    \centering
    \includegraphics[width=0.95\linewidth]{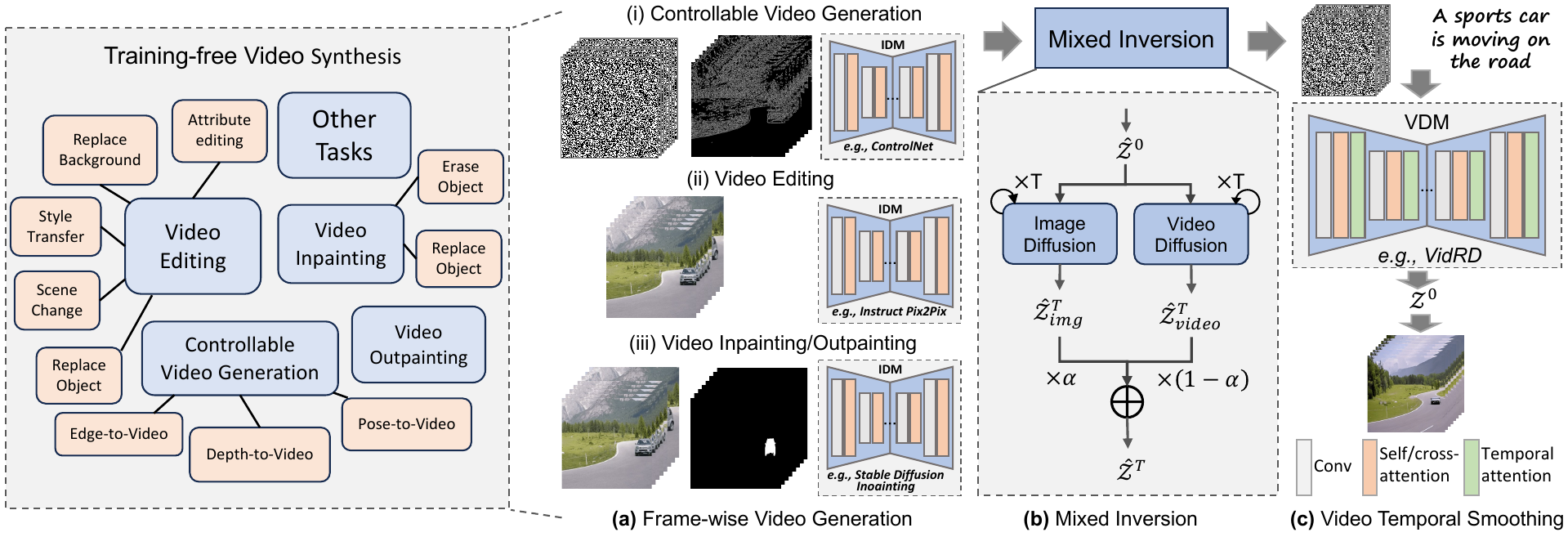}
    \vspace{-2mm}
    \caption{\textbf{BIVDiff pipeline.} Our framework consists of three components, including Frame-wise Video Generation, Mixed Inversion, and Video Temporal Smoothing. We first use the image diffusion model to do frame-wise video generation, then perform Mixed Inversion on the generated video, and finally input the inverted latents into the video diffusion model for video temporal smoothing.}
    \label{fig:framework}
    \vspace{-3mm}
\end{figure*}

\subsection{Diffusion Models for Image Synthesis}
The emergence of diffusion models~\cite{sohl2015deep,ho2020denoising,song2020score} has significantly advanced the progress of text-to-image generation. ADM~\cite{dhariwal2021diffusion} proposes classifier guidance for text-driven image generation. GLIDE~\cite{nichol2021glide} introduces classifier-free guidance~\cite{ho2022classifier} to improve image quality further. DALLE-2~\cite{ramesh2022hierarchical} trains a prior model on CLIP text latents for better text-image alignment. Imagen~\cite{saharia2022photorealistic} shows that text encoding with large language models (e.g., T5~\cite{raffel2020exploring}) is effective at image synthesis. Latent diffusion models (LDM)~\cite{rombach2022high} perform diffusion and denoising processes in latent space, to increase training efficiency.

With the powerful pre-trained text-to-image diffusion foundation models, various downstream image synthesis tasks have also made great progress, such as controllable image generation~\cite{mou2023t2i,zhang2023adding}, image editing~\cite{brooks2023instructpix2pix,DBLP:conf/iclr/HertzMTAPC23,mokady2023null}, personalized image generation~\cite{ruiz2023dreambooth,DBLP:conf/iclr/GalAAPBCC23}, image inpainting~\cite{rombach2022high}, etc. ControlNet~\cite{zhang2023adding} trains an auxiliary U-Net on image-control pairs to make models generate images conditioned on specific controls, such as depth, edge and human pose. Instruct Pix2Pix~\cite{brooks2023instructpix2pix} is trained on generated training data to edit images from instructions. Textual Inversion~\cite{DBLP:conf/iclr/GalAAPBCC23} and DreamBooth~\cite{ruiz2023dreambooth} optimize a single word embedding using a few images of a user-provided concept for personalized image generation. Although effective, additional fine-tuning or optimization on input images is still required to transfer text-to-image foundation models into specific downstream image synthesis tasks, which is costly.

\subsection{Diffusion Models for Video Synthesis}
Inspired by text-to-image diffusion models~\cite{ho2022video,ho2022imagen,DBLP:conf/iclr/SingerPH00ZHYAG23,ge2023preserve,gu2023reuse}, some works propose text-to-video diffusion models by adding extra temporal modules and train models on a large scale of video data. In addition to text-to-video generation, video diffusion models are also applied in various downstream video synthesis tasks, such as controllable video generation~\cite{esser2023structure,chen2023control,xing2023make,wang2023videocomposer} and video editing~\cite{molad2023dreamix,liew2023magicedit}.

Training these video models is memory-hungry and computationally expensive. Some works attempt to adapt pre-trained image diffusion models to videos for efficient video synthesis. Tune-A-Video~\cite{wu2023tune} adopts one-shot tuning on each input video for text-driven video editing. 
VideoP2P\cite{liu2023video} is built on Tune-A-Video~\cite{wu2023tune} and Prompt2Prompt~\cite{DBLP:conf/iclr/HertzMTAPC23}, and introduce Null-Text Inversion~\cite{mokady2023null} to improve the editing quality further. And there are also some training-free video synthesis methods, such as ControlVideo~\cite{zhang2023controlvideo} and FateZero\cite{qi2023fatezero}. ControlVideo~\cite{zhang2023controlvideo} proposes full-frame attention, i.e., concatenating all frames into a "big image" and performing self-attention on it, while Fate-Zero~\cite{qi2023fatezero} fuses self-attention with a blending mask to ensure frame consistency. Although One-shot tuning and optimization make models generate high-fidelity videos, they suffer from poor generalization ability (e.g., cannot edit complex motion). Training-free adapting image diffusion models to videos provides an efficient solution to video synthesis, but they are less effective in maintaining cross-frame consistency at the level of texture and details~\cite{yang2023rerender}, thus flickering artifacts are still severe.

Unlike previous works of adapting image models to videos by adding some modules or complex attention operations for a specific task, we present a simple method to bridge image and video models, and combine both advantages for training-free video synthesis. With a specific downstream image model (e.g., ControlNet~\cite{zhang2023adding}) and a general diffusion-based text-to-video foundation model (e.g., VidRD~\cite{gu2023reuse}), we can efficiently adapt to different video synthesis tasks (e.g., controllable video generation) in a training-free manner.
\section{Method}
\label{sec:framework}

\begin{figure}[t]
    \centering
    \includegraphics[width=1.0\linewidth]{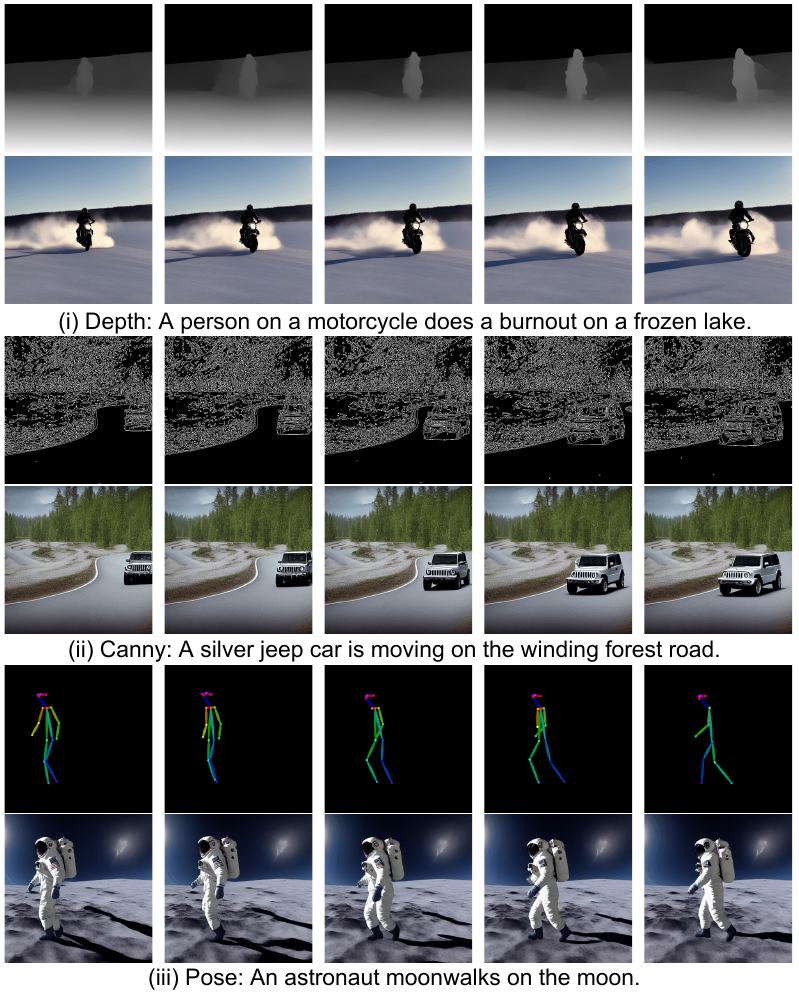}
    \vspace{-7mm}
    \caption{Qualitative results of our proposed BIVDiff on controllable video generation task, conditioned on depth maps, canny edges and human pose sequence. We choose ControlNet~\cite{zhang2023adding} as our image diffusion model.}
    \label{fig:control}
    \vspace{-4mm}
\end{figure}

\begin{figure*}[t]
    \centering
    \includegraphics[width=0.85\linewidth]{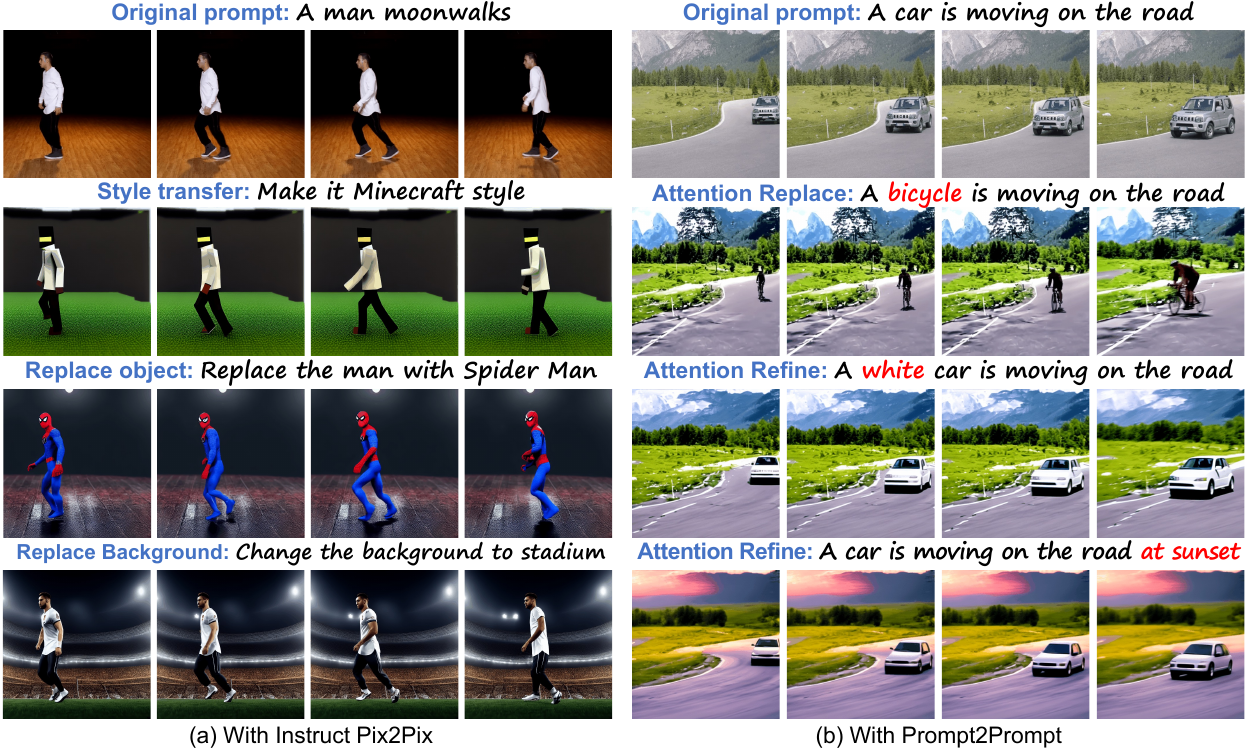}
    \vspace{-2.5mm}
    \caption{Qualitative results of our proposed BIVDiff on video editing task. We select two popular image editing methods, Instruct Pix2Pix~\cite{brooks2023instructpix2pix} and Prompt2Prompt~\cite{DBLP:conf/iclr/HertzMTAPC23} as image models, and test a wide range of editing types.}
    \label{fig:edit}
    \vspace{-3mm}
\end{figure*}

Given a video synthesis task, we choose an image diffusion model \textbf{(IDM)} of its image task version and a text-to-video diffusion foundation model \textbf{(VDM)}. Let random latents $\mathcal{Z}^T = \{z_i^T\}_{i=1}^m$ or video $\mathcal{V} = \{v_i\}_{i=1}^m$ be the inputs, where $T$ is the number of diffusion step, and $m$ is frame number. Let $\mathcal{P}^\ast$ be the target prompt, and $\mathcal{C}$ be the conditions (e.g., depth maps and masks) according to target task. Our goal is to generate a temporally coherent video $\mathcal{V}^\ast$.

Our framework consists of three components, including Frame-wise Video Generation, Mixed Inversion, and Video Temporal Smoothing. As shown in \cref{fig:framework}, we first use the image diffusion model to perform frame-wise video generation, then perform Mixed Inversion on the generated video, and finally input the inverted latents into the video diffusion model for video temporal smoothing. 





\subsection{Frame-wise Video Generation}
\label{subsec:frames}
The first step of our proposed framework is to perform frame-wise video generation with image diffusion models. For the given video synthesis task, we can choose an image counterpart. For instance, if we want to do controllable video generation, then we can use ControlNet~\cite{zhang2023adding} to generate the frames under the conditioning controls (e.g., edges, depth, etc.) independently. As for video editing, we can select one image editing model, such as Instruct Pix2Pix~\cite{brooks2023instructpix2pix}, to edit each frame in the video according to the target prompt independently. The generation process can be formulated as:
\begin{equation}
    \hat{\mathcal{Z}^0} = \{\hat{z}_i^0 = \text{IDM}(f_i, \mathcal{C})\}_{i=1}^m,
\end{equation}
where $f_i$ is the $i$-th random latent or frame in the given video, and $\mathcal{C}$ are conditions (e.g., text prompt, depth maps, and masks). Due to this decoupled design, our framework gains great flexibility and strong generalization ability. That is to say that we can choose arbitrary downstream image diffusion models for general-purpose video synthesis.

\subsection{Mixed Inversion}
\label{subsec:inversion}

The key of bridging image and video diffusion models is DDIM Inversion. After IDM denoising, we need to conduct DDIM Inversion to convert denoised latents to initial noisy latents as the input to the subsequent VDM. By DDIM Inversion, we can preserve the information that IDM generates, and make VDM synthesized videos consistent with the results of IDM but temporally coherent, instead of free generation. The frame-wise DDIM Inversion process can be formulated as:
\[
    \hat{\mathcal{Z}^T} = \{\hat{z}_i^T = \text{DDIM}_{\text{inv}}^{\text{img}}(\hat{z}_i^0)\}_{i=1}^m, 
\]
where $\text{DDIM}_{\text{inv}}^{\text{img}}$ means DDIM Inversion with image diffusion models. It is worth noting that we choose an image diffusion foundation model (e.g., Stable Diffusion~\cite{rombach2022high}) for DDIM Inversion instead of the same model for frame-wise video generation, and the prompt is $\phi$ for DDIM Inversion.

Despite using inverted latents by image DDIM Inversion, VDM tends to generate content inconsistent with IDM in some cases, due to the distribution shifts. Moreover, when the gap between the latent distributions of image and video diffusion models is big, VDMs will fail to generate correct videos. For example, in the cases of inputting source videos, the initial noised latents of frames obtained by frame-wise DDIM Inversion with image diffusion models are highly correlated, making some VDMs (e.g., VidRD~\cite{gu2023reuse}) requiring i.i.d. random latents as inputs collapse and generate meaningless noises (\cref{fig:ablation_mixing_ratio}). 

To solve these problems, we introduce Mixed Inversion. As shown in \cref{fig:framework}, we perform DDIM Inversion with both image and video diffusion models. Both latents by Image and Video DDIM Inversion keep the contents of videos. The former can be temporally smoothed by VDM but the distribution may be different from the distributions required by VDM. The latter cannot be further temporally smoothed by VDM distribution but the distribution is consistent with VDM. We can weighted-sum these two latents to adjust the distribution of initial latents fed into VDM. The latents mixing process is as follows:
\begin{align}
    &\hat{\mathcal{Z}}_{img}^T = \{\hat{z_i^T} = \text{DDIM}_{\text{inv}}^{\text{img}}(\hat{z_i^0})\}_{i=1}^m, \\
    &\hat{\mathcal{Z}}_{video}^T = \text{DDIM}_{\text{inv}}^{\text{video}}(\hat{\mathcal{Z}^0}),\\
    &\hat{\mathcal{Z}^T} = \alpha \cdot \hat{\mathcal{Z}}_{img}^T + (1-\alpha) \cdot \hat{\mathcal{Z}}_{video}^T,
\end{align}
where $\text{DDIM}_{\text{inv}}^{\text{video}}$ means DDIM Inversion with our video diffusion model~\cite{gu2023reuse} and $\alpha$ is the mixing ratio used to adjust the ratio of the image and video latent components.
With Mixed Inversion, we can adjust the latent distribution to make VDMs produce correct results, and trade eoff between temporal smoothing and open generation capability of VDMs. 

\begin{figure*}[t]
    \centering
    \includegraphics[width=0.9\linewidth]{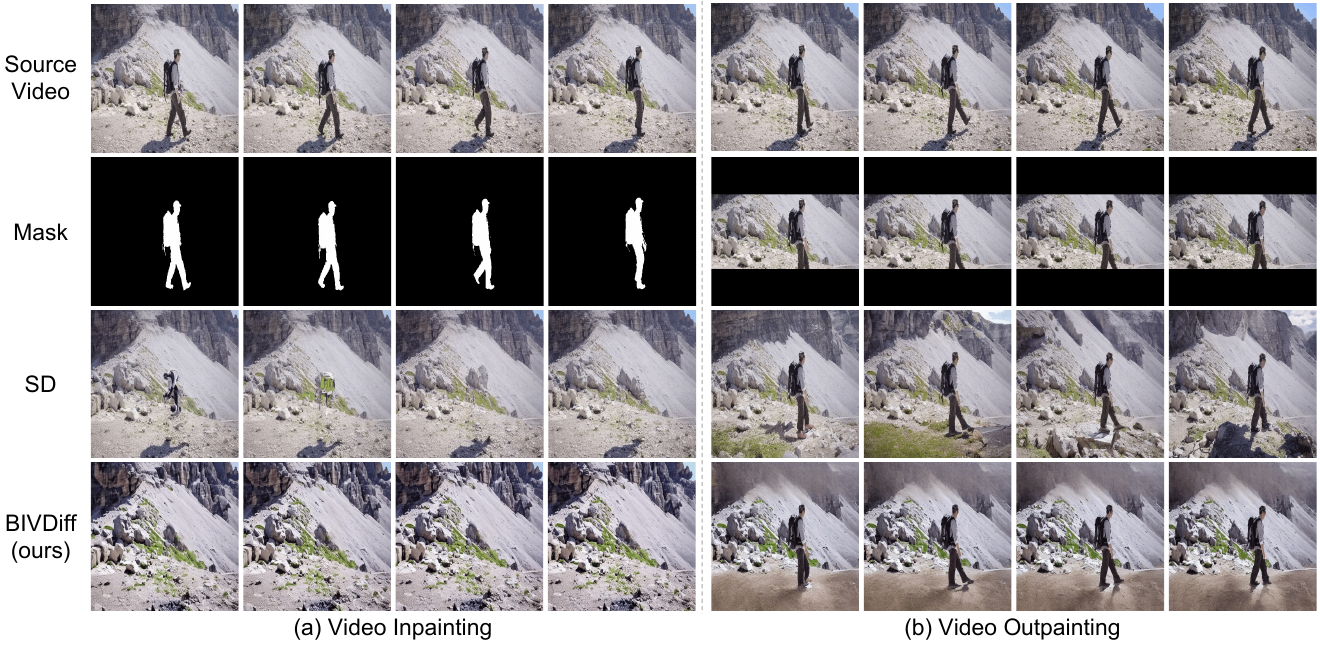}
    \vspace{-2.5mm}
    \caption{Qualitative results of our proposed BIVDiff on video inpainting and outpainting task. We adopt Stable Diffusion Inpainting~\cite{rombach2022high} as our image model. Our method can erase objects and complete the masked regions well.}
    \label{fig:inpainting_and_outpainting}
    \vspace{-3mm}
\end{figure*}

\subsection{Video Temporal Smoothing}
\label{subsec:smoothing}
Although we can resort to image diffusion models for video synthesis tasks by frame-wise generation, temporal consistency is ignored, leading to visible flickers (e.g., \cref{fig:compairison}). Video generation foundation models learn temporal consistency and can generate temporally coherent videos. Therefore, we do temporal smoothing on the video generated by IDM, by feeding the inverted latents into VDM. VDM can effectively capture the information stored in the inverted latents, and make the input videos consistent in temporal dimension, without destroying the contents created by IDM. The temporal smoothing process is formulated as:
\begin{equation}
    \mathcal{Z}^0 = \text{VDM}(\hat{\mathcal{Z}^T}, \mathcal{P}^\ast).
\end{equation}
After temporal smoothing, we use vae decoder to decode the denoised latents to the target video.

\section{Experiment}
\label{sec:experiment}

\subsection{Implementation Details}
To validate the effectiveness of our framework, we perform experiments on four representative video synthesis tasks, including 1) controllable video generation with ControlNet~\cite{zhang2023adding}, 2) video editing with Instruct Pix2Pix~\cite{brooks2023instructpix2pix} and Prompt2Prompt~\cite{DBLP:conf/iclr/HertzMTAPC23}, 3) video inpainting with Stable Diffusion Inpainting~\cite{rombach2022high} and 4) video outpainting with Stable Diffusion Inpainting~\cite{rombach2022high}. For the video diffusion foundation model, we choose VidRD~\cite{gu2023reuse}. In the case of models for DDIM Inversion, we use Stable Diffusion 1.5 for frame-wise inversion, and VidRD for video-level inversion.

In our experiments, we generate 8 frames with 512 $\times$ 512 resolution for each video. The classifier-free guidance scale is 7.5 and the total timestep is 50. For the mixing ratio $\alpha$ in Mixed Inversion, we set 1.0, 1.0, 0.25, and 0.1 for BIVDiff with ControlNet, Instruct Pix2Pix, Prompt2Prompt and Stable Diffusion Inpainting as the default settings, respectively. And there is no per-video optimization (e.g. Null-text Inversion~\cite{mokady2023null}) in our experiments.

\subsection{Qualitative Results}
\textbf{Controllable Video Generation.} By bridging pre-trained controllable image generation model ControlNet~\cite{zhang2023adding} and text-to-video foundation model VidRD~\cite{gu2023reuse}, our framework BIVDiff supports zero-shot controllable video generation. \cref{fig:control} shows the generated videos conditioned on depth maps, canny edge maps, and human pose sequences. As shown in \cref{fig:control}, the generated videos are well-matched with the conditions and keep significant temporal consistency, such as backgrounds, and both the appearance and structure of foreground objects.

\noindent \textbf{Video Editing.} Video editing is another important application in video synthesis. We choose two representative image editing models Instruct Pix2Pix~\cite{brooks2023instructpix2pix} and Prompt2Prompt~\cite{DBLP:conf/iclr/HertzMTAPC23} for zero-shot video editing. For video editing with Instruct Pix2Pix, we test various editing types, including style transfer, object replacement, and background replacement, as shown in \cref{fig:edit} (a). As for Prompt2Prompt, we follow the paper to do attention replacement and attention refinement. As shown in \cref{fig:edit} (b), our framework can replace the object, edit attribute, and do global editing, which are inherited from Prompt2Prompt in a zero-shot manner. 

\noindent \textbf{Video Inpainting and Outpainting.} Additionally, we introduce an image inpainting model Stable Diffusion Inpainting~\cite{rombach2022high} for video inpainting. For video outpainting, we can transfer the inpainting model to outpainting easily, by making masked regions of outpainting be the erased regions of inpainting. As shown in \cref{fig:inpainting_and_outpainting}, independently processing each frame makes imperfect shadows that have not been completely erased and inconsistent areas to be filled in. We can eliminate these temporal inconsistencies by combining image and video diffusion models.

\noindent \textbf{Additional Models.} To further validate the effectiveness and general use of BIVDiff, we introduce more diffusion models, including another video diffusion model ZeroScope~\cite{cerspense2023zeroscope} and image diffusion model T2I-Adapter~\cite{mou2023t2i}. The qualitative results are in Supplementary Material.

\begin{figure*}[t]
    \centering
    \includegraphics[width=1.0\linewidth]{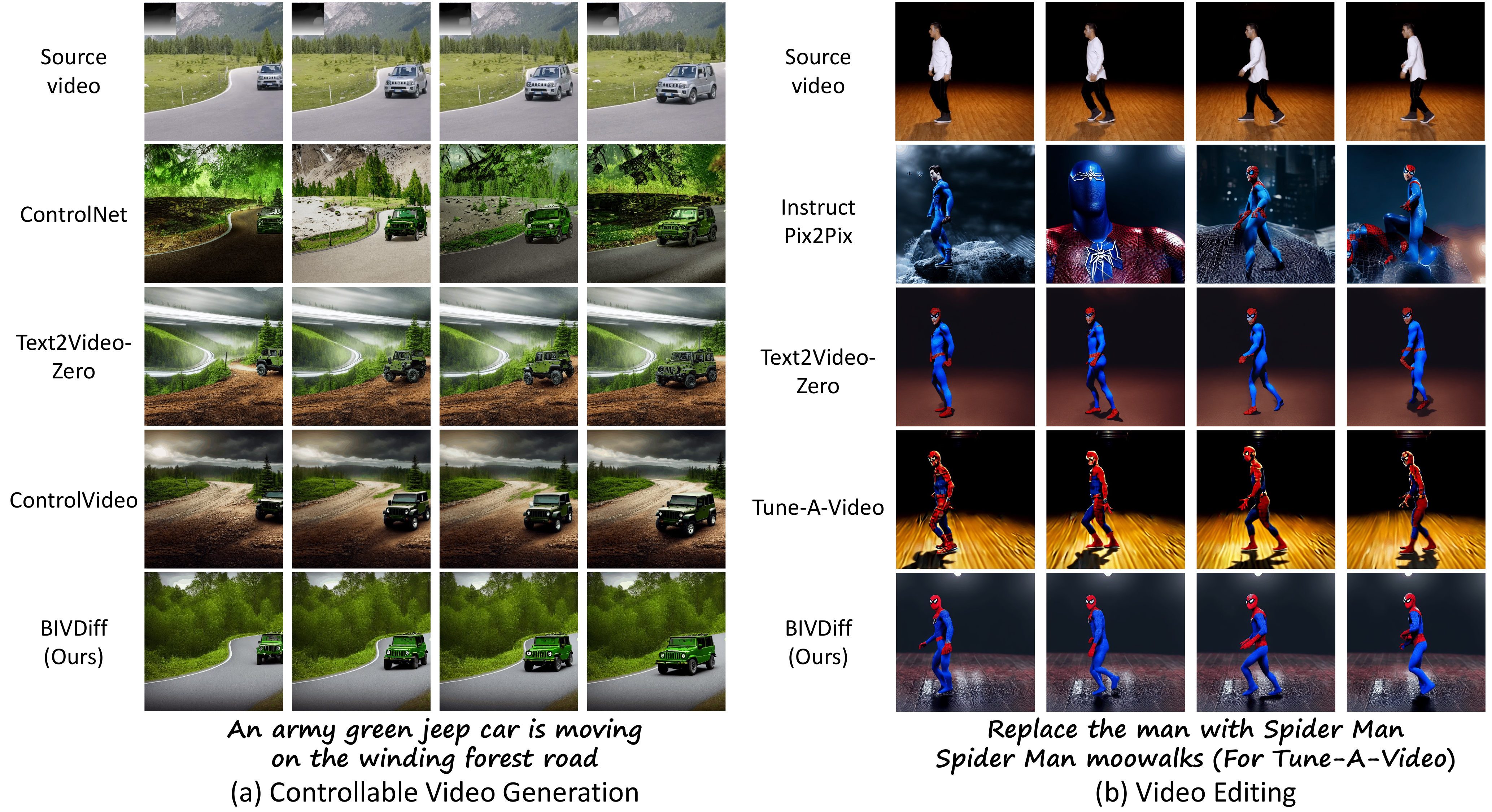}
    \vspace{-7mm}
    \caption{Qualitative comparison with baselines on controllable video generation and video editing task. Our BIVDiff generates high-quality and temporally coherent videos, and shows better (a) control and temporal consistency and (b) fidelity and realness.}
    \label{fig:compairison}
\end{figure*}

\begin{table*}[t]
\center
\resizebox{.75\textwidth}{!}{
\begin{tabular}{lccccccccc}
        \hline
        \multirow{2}{*}{Method} & \multicolumn{2}{c}{Automatic Metrics} &  & \multicolumn{4}{c}{User Study}   & & Inference Time                          \\ \cline{2-3} \cline{5-8} 
                                & Frame Consistency & Textual Alignment &  & Quality       & Alignment     & Fidelity      & Avg.  & &(per video)        \\ \hline
        Text2Video-Zero         & 91.69             & \textbf{26.85}             &  & 2.74          & 3.16          & \textbf{2.98} & 2.96    & &  \textbf{25s}    \\
        ControlVideo           & 92.63             & 26.12             &  & 2.61          & 3.12          & 2.54          & 2.76         & & 57s \\
        BIVDiff (Ours)          & \textbf{92.67}             & 26.25             &  & \textbf{3.38} & \textbf{3.24} & 2.72          & \textbf{3.11} & & 61s \\ \hline
        Text2Video-Zero        & 91.57             & 25.37             &  & 2.26          & 2.23          & 2.46 & 2.32      & & \textbf{56s}   \\
        FateZero           & 90.75             & 26.42             &  & 2.38          & 1.7          & \textbf{3.05}          & 2.38    & & 221s     \\
        Tune-A-Video            & 90.46             & \textbf{28.33}             &  & 2.30          & 2.23          & 2.35          & 2.29          & & 11min + 26s\\
        BIVDiff (Ours)          & \textbf{93.50}             & 26.16             &  & \textbf{2.98} & \textbf{2.30} & 2.68          & \textbf{2.65} & & 64s\\ \hline
    \end{tabular}
}
\vspace{-2mm}
\caption{Quantitative comparison with baselines. The upper part is the result of controllable video generation with depth control. The bottom part is the result of video editing. Tune-A-Video adopts null-text inversion and one-shot tuning, while Text2Video-Zero and our BIVDiff are based on InstructPix2Pix and training-free. Our method has the best temporal consistency and is most favored by humans.}
\label{tab:comparison}
\vspace{-5mm}
\end{table*}

\begin{figure}[t]
    \centering
    \includegraphics[width=0.75\linewidth]{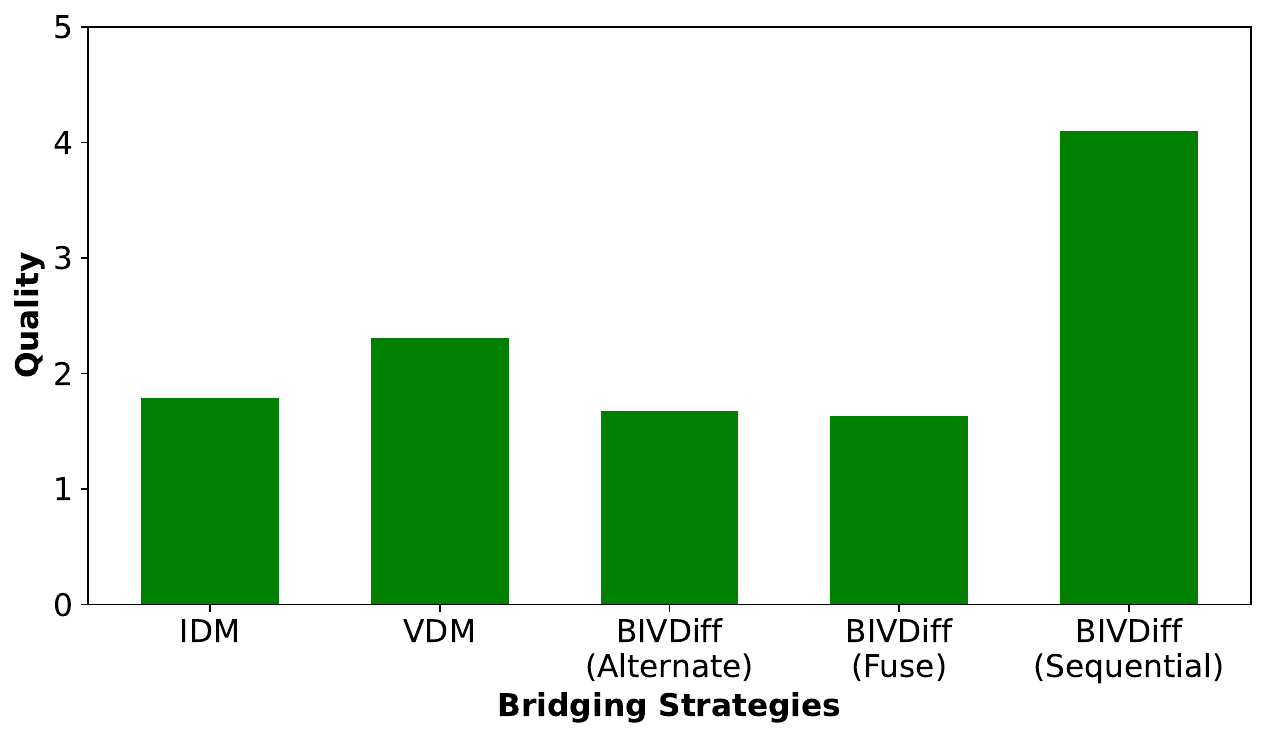}
    \vspace{-4mm}
    \captionof{figure}{User study for bridging strategies ablation study.}
    \label{fig:abaltion_bridging_strategies_user_study}
    \vspace{-5mm}
\end{figure}

\begin{figure}[t]
    \centering
    \includegraphics[width=0.78\linewidth]{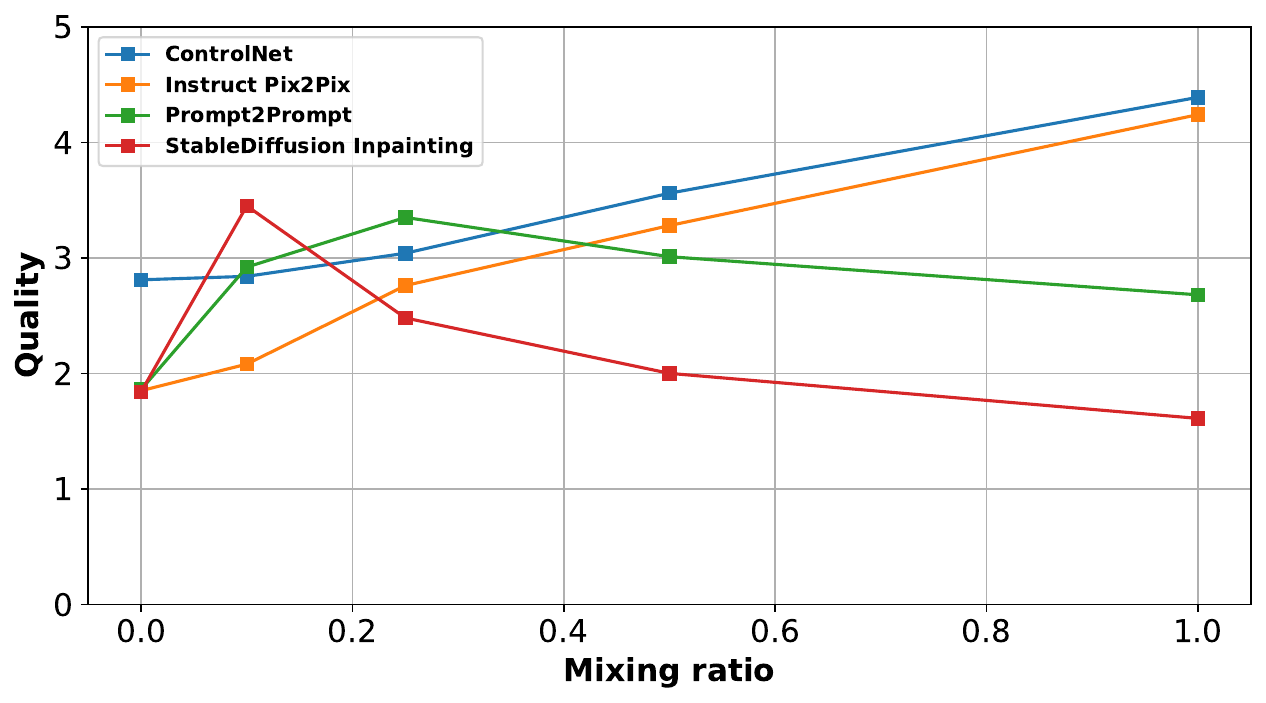}
    \vspace{-4mm}
    \captionof{figure}{User study for mixing ratio ablation study.}
    \label{fig:abaltion_mixing_ratio_user_study}
    \vspace{-5mm}
\end{figure}

\begin{figure}[t]
    \centering
    \includegraphics[width=0.95\linewidth]{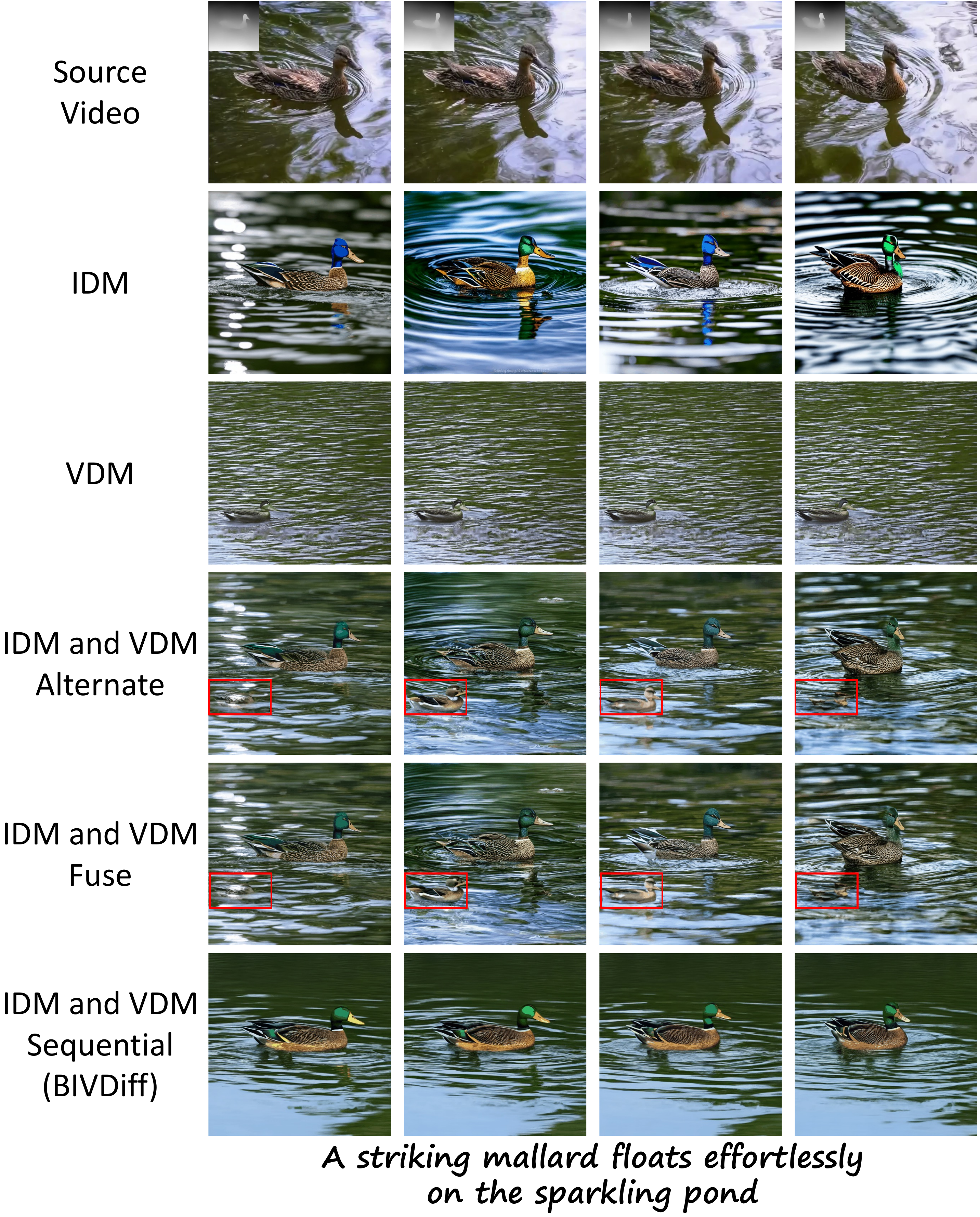}
    \vspace{-3mm}
    \caption{Ablation on strategies of bridging image and video diffusion models. Our sequential strategy can temporally smooth the videos (e.g., consistent appearance, structure, and background), and limit the open generation ability of VDM (e.g., the generated mallard of VDM is not in the final result of our method.)}
    \label{fig:ablation_framework}
    \vspace{-6mm}
\end{figure}

\begin{figure}[t]
    \centering
    \includegraphics[width=0.95\linewidth]{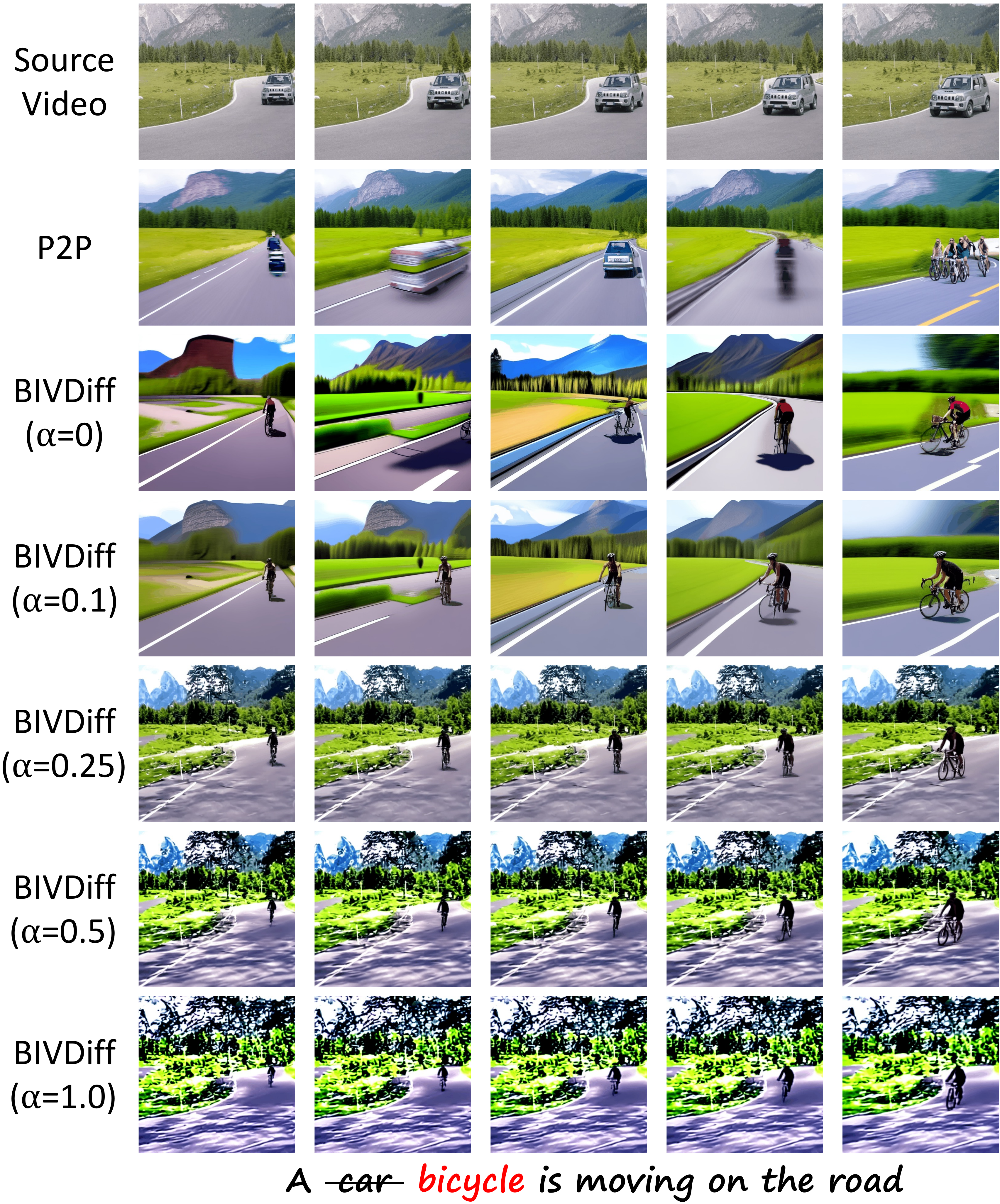}
    \vspace{-2.5mm}
    \caption{Ablation on the mixing ratio $\alpha$ in Mixed Inversion. Larger $\alpha$ leads to more temporally consistent videos, and smaller $\alpha$ makes the distribution of latents fed into VDM closer to VDM's and generates higher quality videos.}
    \label{fig:ablation_mixing_ratio}
    \vspace{-5mm}
\end{figure}

\subsection{Comparison with Baselines}
We quantitatively and qualitatively compare our method with some baselines on controllable video generation (Text2Video-Zero~\cite{khachatryan2023text2video}, FateZero~\cite{qi2023fatezero} and Tune-A-Video\cite{zhang2023controlvideo}) and video editing (Text2Video-Zero~\cite{khachatryan2023text2video} and ControlVideo~\cite{khachatryan2023text2video}). For quantitative comparison, we use DAVIS dataset in LOVEU-TGVE Benchmark~\cite{wu2023cvpr}, which consists of 16 videos and with 4 prompts per video, for automatic metrics and user study evaluation. Following Tune-A-Video~\cite{wu2023tune}, we adopt CLIP~\cite{radford2021learning} to calculate frame consistency and textural alignment score. For user study, we follow Dreamix~\cite{molad2023dreamix} to invite 25 human raters working on AI, arts and other areas, to rate videos by quality, fidelity, and alignment score on a scale of $1-5$. We also test the practical running time to compare inference speed.

\noindent \textbf{Quantitative Comparison.} Table \ref{tab:comparison} shows the quantitative results. For automatic metrics, our method has the best frame consistency due to the strong temporal modeling of VDM and comparable textual alignment. And our method is most favored by participants in the user study experiment since we can generate temporally coherent and realistic high-quality videos. Moreover, BIVDiff achieves a comparable inference speed in practice. Without modifying structures and inference pipelines inside IDM and VDM, we avoid time-consuming attention operations~\cite{qi2023fatezero} or training~\cite{wu2023tune} and benefit from parallel GPU computing.

\noindent \textbf{Qualitative Comparison.} We present visual comparisons in \cref{fig:compairison}. \cref{fig:compairison}(a) shows that ControlNet generates high-quality frames matched with controls (e.g., depth maps), but has severe frame inconsistency (e.g., the background is inconsistent across frames). Text2Video-Zero and ControlVideo generate temporally smooth videos, but there are still some slight flickers due to weak temporal modeling, and they struggle to accurately match the given controls (e.g., the lane lines disappear). In contrast, our method can generate temporally coherent videos well-matched with the conditions. Similar results can be found in video editing (\cref{fig:compairison}(b)). Our method can keep more details in the input video (e.g., floor and shadows) and the generated videos are more realistic (e.g., the body of Spider Man).

\subsection{Ablation Study}
In this section, we study several key designs of our method, including the strategies of bridging image and video diffusion models, and the mixing ratio $\alpha$ in Mixed Inversion.

\noindent \textbf{Ablation on bridging strategies.} To validate the effectiveness of our bridging framework, we realize different strategies for comparisons, including 1) \textbf{IDM}. We use ControlNet~\cite{zhang2023adding} to do frame-wise video generation under the guidance of depth control. 2) \textbf{VDM}. We adopt VidRD~\cite{gu2023reuse} for text-to-video generation without depth control. 3) \textbf{IDM and VDM Alternate}. We use IDM and VDM for alternate denoising, i.e. one IDM denoising step by one VDM denoising step. 4) \textbf{IDM and VDM Fuse}. We perform IDM and VDM denoising simultaneously, and average these two latents. 5) \textbf{IDM and VDM Sequential}, i.e. our proposed BIVDiff. 
The user study in \cref{fig:abaltion_bridging_strategies_user_study} shows our sequential strategy works best. As shown in \cref{fig:ablation_framework}, videos generated by ControlNet are temporally inconsistent and VDM produces temporally consistent videos but unmatched with the given depth control. Bridging IDM and VDM during the denoising process (``Alternate'' and ``Fuse'') tries to combine the results of IDM and VDM (e.g., there are two mallards in the videos). In contrast, our proposed BIVDiff bridges IDM and VDM in a sequential way, and generates temporally coherent videos consistent with depth control.

\noindent \textbf{Ablation on mixing ratios.} 
 We also conduct an ablation study on video editing with Prompt2Prompt~\cite{DBLP:conf/iclr/HertzMTAPC23} to analyze the effects of mixing ratio. As shown in \cref{fig:ablation_mixing_ratio}, the larger $\alpha$ is, the more temporally consistent the generated videos are. For example, there is a car and multiple bicycles that should not have appeared in the edited videos of Prompt2Prompt~\cite{DBLP:conf/iclr/HertzMTAPC23}. In contrast, videos generated by our method are more consistent with the input video and text prompt, and temporally coherent when $\alpha$ is 0.25. However, with $\alpha$ increasing, the quality of synthesized videos degrades and there are a lot of noises and artifacts in the videos. This is because the frames in the edited videos are similar (e.g., similar large areas of background) and latents by frame-wise DDIM Inversion with image diffusion models are highly correlated. When the video diffusion models, such as VidRD~\cite{gu2023reuse}, require i.i.d. random latents as input, models will corrupt and produce noised videos. \cref{fig:abaltion_mixing_ratio_user_study} shows video quality under different mixing ratios for each IDM and VDM pair. In practice, we can use small $\alpha$ to bridge latent distribution gaps and generate correct videos.

\section{Conclusion}
\label{sec:conclusion}
In this paper, we present a training-free framework for general-purpose video synthesis, coined as BIVDiff, via bridging downstream image diffusion models and text-to-video foundation diffusion models. We first use an image diffusion model (e.g., ControlNet~\cite{zhang2023adding}) for frame-wise video generation, then perform Mixed Inversion on the generated video, and finally input the inverted latents into the video diffusion model (e.g., ViDRD~\cite{gu2023reuse}) for temporal smoothing. We introduce Mixed Inversion to adjust the latent distribution to make VDMs produce correct results, and balance between temporal smoothing and open generation capability of VDMs. Extensive experiments on a wide range of video synthesis tasks demonstrate the effectiveness and generalization power of our method.

\noindent
{\bf Acknowledgements.} This work is supported by National Key R$\&$D Program of China (No. 2022ZD0160900), National Natural Science Foundation of China (No. 62076119, No. 61921006)), and Collaborative Innovation Center of Novel Software Technology and Industrialization.

\clearpage
{
    \small
    \bibliographystyle{ieeenat_fullname}
    \bibliography{main}
}

\maketitlesupplementary

\section{More models}
\label{sec:more_models}

To further validate the effectiveness and general use of BIVDiff, we introduce more diffusion models into our proposed BIVDiff framework.

\begin{figure*}[t]
    \centering
    \includegraphics[width=1.0\linewidth]{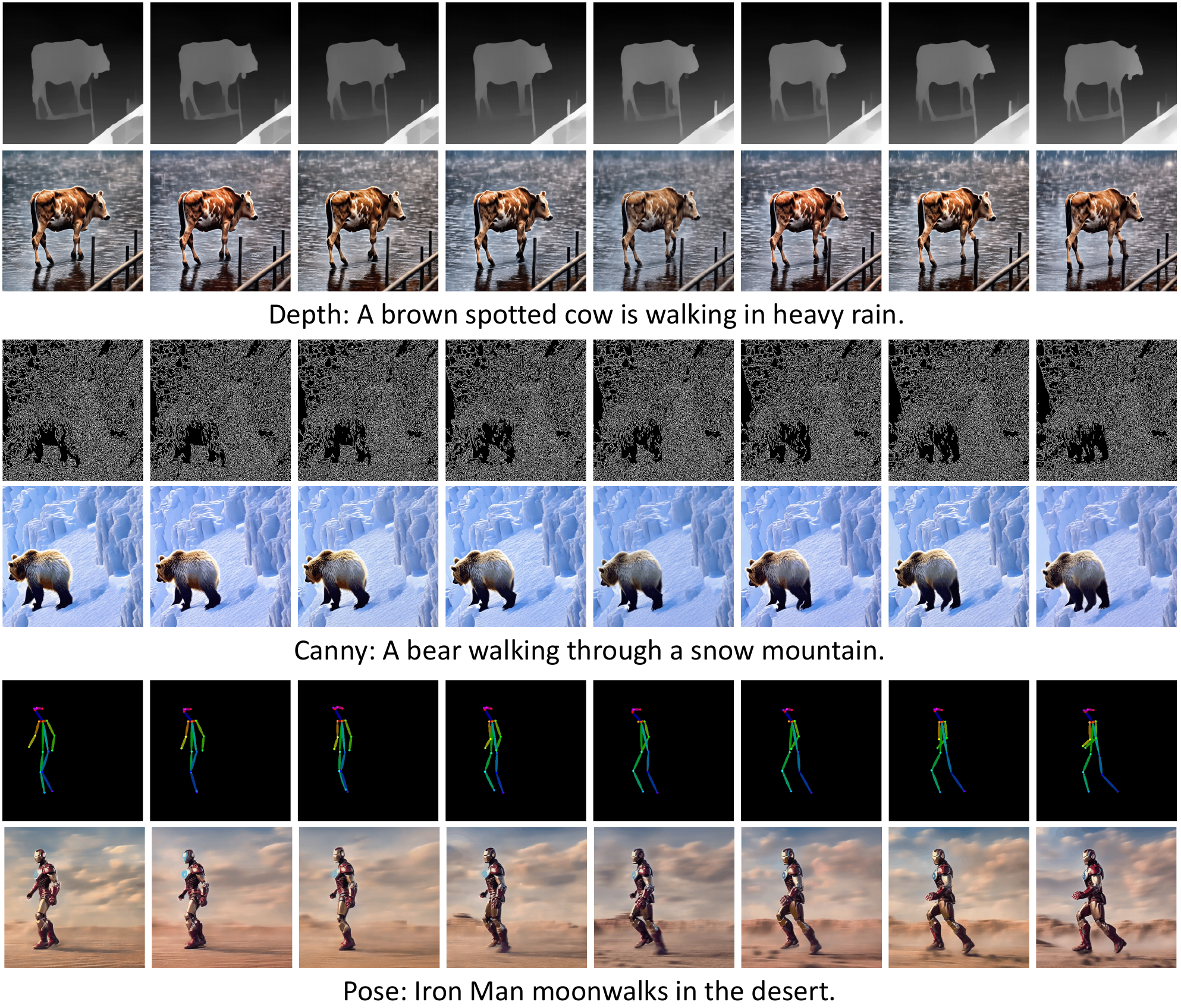}
    \caption{Qualitative results of our proposed BIVDiff on controllable video generation task, conditioned on depth maps, canny edges and human pose sequence. We choose ControlNet~\cite{zhang2023adding} as our image diffusion model and ZeroScope as our video diffusion model.}
    \label{fig:zeroscope_control}
\end{figure*}

\begin{figure*}[t]
    \centering
    \includegraphics[width=1.0\linewidth]{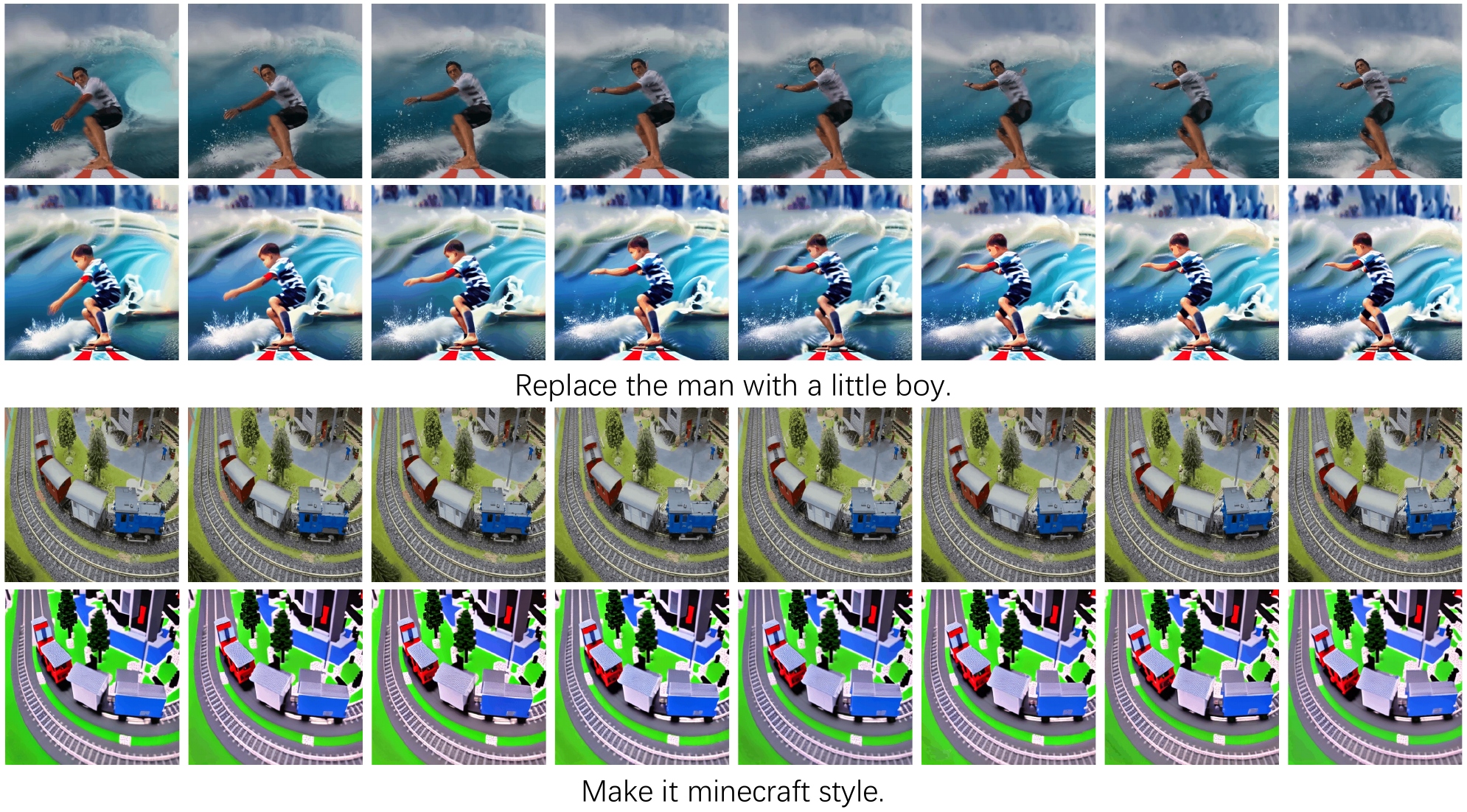}
    \caption{Qualitative results of our proposed BIVDiff on video editing task. We choose Instruct Pix2Pix~\cite{brooks2023instructpix2pix} and ZeroScope as our video diffusion model.}
    \label{fig:zeroscope_edit}
\end{figure*}

\noindent \textbf{Additional Video Model.} In addition to VidRD~\cite{gu2023reuse}, we use another video diffusion foundation model ZeroScope~\cite{cerspense2023zeroscope} as our VDM to perform video temporal smoothing. Specifically, we perform controllable video generation with ControlNet~\cite{zhang2023adding} conditioned on depth maps, canny edge maps and human pose sequence, and video editing task with Instruct Pix2Pix~\cite{brooks2023instructpix2pix}. As shown in \cref{fig:zeroscope_control} and \cref{fig:zeroscope_edit}, the generated videos keep temporal consistency well, demonstrating the flexity of model selections and general use of our BIVDiff framework. 

\begin{figure*}[t]
    \centering
    \includegraphics[width=1.0\linewidth]{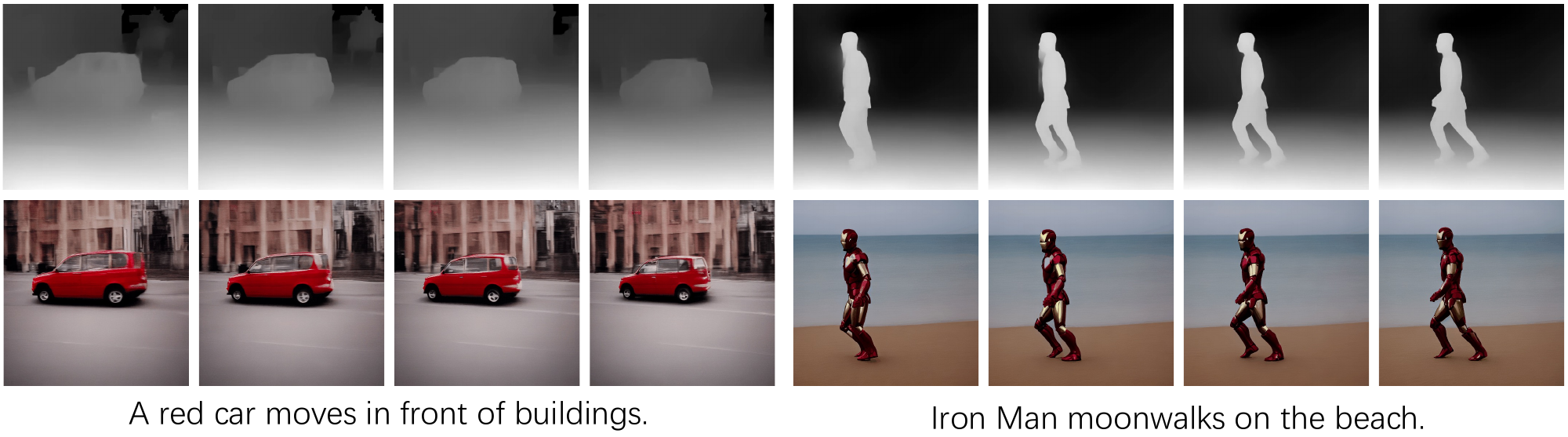}
    \caption{Qualitative results of our proposed BIVDiff on controllable video generation task, conditioned on depth maps. We choose T2I-Adapter~\cite{mou2023t2i} as our image diffusion model and VidRD~\cite{gu2023reuse} as our video diffusion model.}
    \label{fig:t2i_adapter}
\end{figure*}

\noindent \textbf{Additional Image Model.} We choose another popular controllable image generation model T2I-Adapter~\cite{mou2023t2i} as our IDM for controllable video generation, conditioned on depth maps. As shown in \cref{fig:t2i_adapter}, the generated videos keep temporal consistency well and are consistent with the given controls, demonstrating the flexity of model selections and general use of our BIVDiff framework. 

\begin{figure*}[t]
    \centering
    \includegraphics[width=1.0\linewidth]{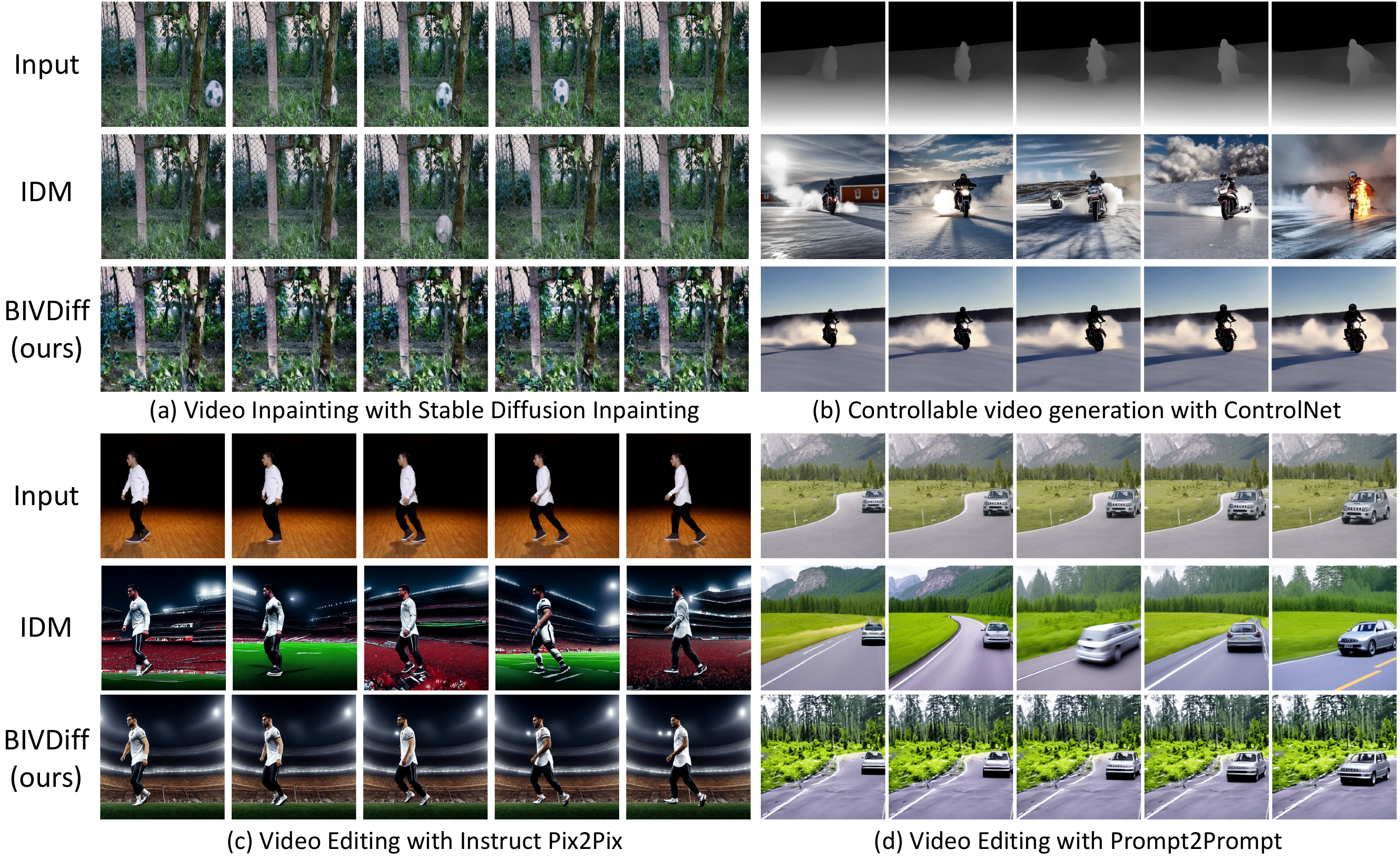}
    \caption{\textbf{Effects of Video Temporal Smoothing.} We compare IDM (using image models only) and our proposed BIVDiff (bridging image and video models), to validate the effectiveness of temporal smoothing power brought by VDM.}
    \label{fig:temporal_smoothing}
\end{figure*}

\begin{figure*}[t]
    \centering
    \includegraphics[width=1.0\linewidth]{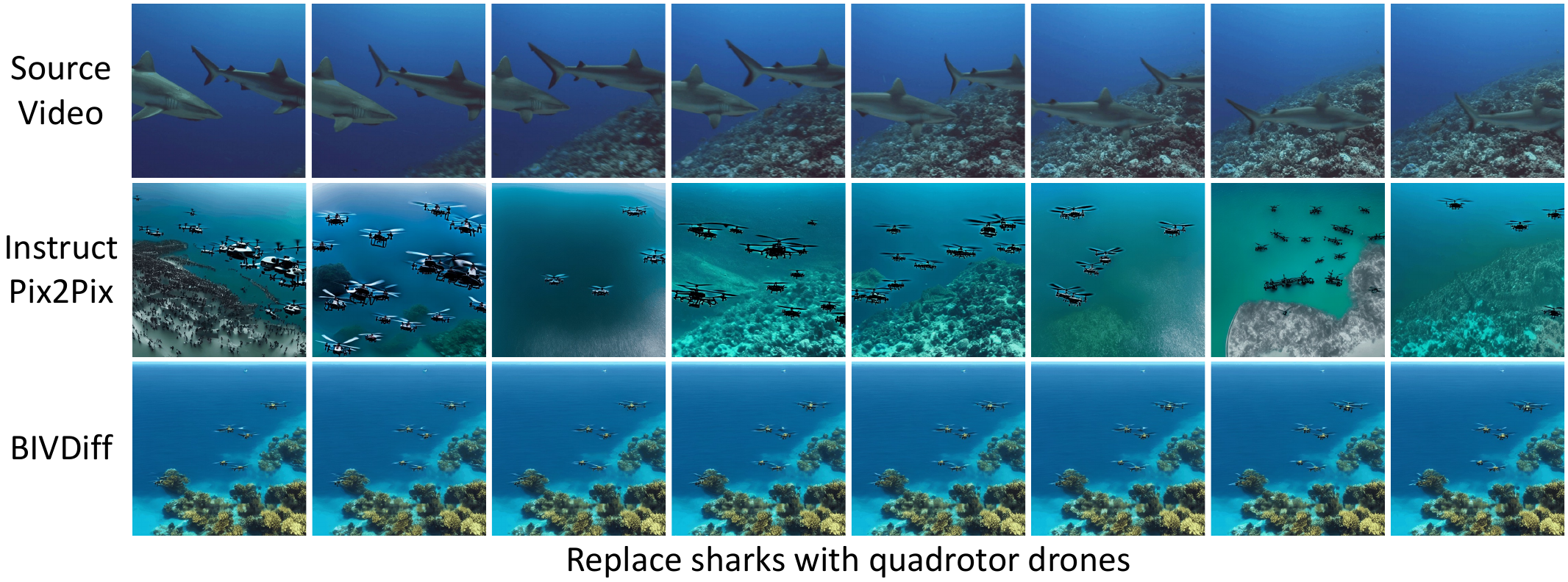}
    \caption{\textbf{A failure case of video editing.} Our method may produce unsatisfied results when the results get by frame-wise video generation with only image models are far away from expectations.}
    \label{fig:failure_cases}
\end{figure*}

\section{Effects of Video Temporal Smoothing}
In \cref{fig:temporal_smoothing}, we show several cases to compare using image diffusion models (IDM) for frame-wise generation and our proposed BIVDiff which bridges image and video diffusion models, to validate the effectiveness of video temporal smoothing provided by VDM. Using IDM only produces temporally inconsistent videos for lacking temporal modeling. By bridging task-specific image models and video diffusion foundation models, we can produce temporally coherent videos (e.g., consistent appearance and structure of foreground objects and background across frames), while performing the target task well (e.g., well-matched with given controls for controllable video generation, and keeping good fidelity for video editing).

\section{User Study Details}
\label{sec:user_study}
Following Dreamix~\cite{molad2023dreamix}, we invite 25 human raters working on AI, arts and other areas, to rate videos by quality, alignment, and fidelity on a scale of $1-5$ (1 is the lowest score and 5 is the highest score). The explanations of these metrics are as follows:

\begin{enumerate}
    \item \textbf{Quality:} Rate the overall visual quality and smoothness of the edited
video.
    \item \textbf{Alignment:} How well does the edited video match the textual edit description provided?
    \item \textbf{Fidelity:} How well does the edited video preserve unedited details of
the original video?
\end{enumerate}

For quantitative comparisons, we perform user study on DAVIS dataset in LOVEU-TGVE Benchmark~\cite{wu2023cvpr}, which consists of 16 videos and with 4 prompts per video. For ablation studies, we evaluate bridging strategies on four model pairs (ContorlNet and InstructPix2Pix as IDMs, and VidRD and ZeroScope as VDMs) with 10 videos and 16 text prompts. For mixing ratio, we use 8 videos, and test four IDMs (VidRD as VDM) with 5 prompts for each IDM.

\section{Limitations}
\cref{fig:failure_cases} shows a failure case of our method, where the edited video is inconsistent with the input video (i.e., low fidelity). This is due to the wrong editing results of Instruct Pix2Pix. When the results of frame-wise video generation with only image models are far away from expectations, our method may produce unsatisfied results. Luckily, due to the flexible image model selection brought by decoupling image and video models in our framework, we can tackle this problem by simply choosing another image diffusion model to generate correct results.

\end{document}